\documentclass[letterpaper]{article} 
\usepackage{aaai24}      
\usepackage{times}                   
\usepackage{helvet}                  
\usepackage{courier}                 
\usepackage[hyphens]{url}            
\usepackage{graphicx}                
\usepackage{natbib}                  
\usepackage{todonotes}
\usepackage{caption}                 
\usepackage{algorithm}
\usepackage{algorithmicx}
\usepackage{ dsfont }

\usepackage{newfloat}
\usepackage{listings}
\DeclareCaptionStyle{ruled}{labelfont=normalfont,labelsep=colon,strut=off} 
\floatstyle{ruled}
\newfloat{listing}{tb}{lst}{}
\floatname{listing}{Listing}

\usepackage[utf8]{inputenc} 
\usepackage[T1]{fontenc}    
\usepackage{url} 
\usepackage{booktabs}       
\usepackage{amsfonts}       
\usepackage{nicefrac}       
\usepackage{microtype}      
\usepackage{xcolor}         
\usepackage{mathtools}
\usepackage[noend]{algpseudocode}
\usepackage{todonotes}
\usepackage{amsthm}
\usepackage{adjustbox}
\usepackage{todonotes}
\usepackage{amssymb}
\usepackage{comment}
\usepackage{caption}
\usepackage{subcaption}
\usepackage{xspace}
\usepackage{booktabs}
\usepackage{tabularx}
\usepackage{multirow}
\usepackage{amsmath}

\DeclareMathOperator*{\argmax}{arg\,max}

\usepackage{tikz}
\usetikzlibrary{shapes.geometric}
\usetikzlibrary{positioning,automata,arrows}
\usepackage{pifont}
\usepackage{fontawesome,wasysym,marvosym}

\newcommand{\coffee}[0]{{\color{black}\Coffeecup}\xspace}
\newcommand{\mail}[0]{\Letter}

\newcommand{\agent}{\resizebox{4mm}{!}{\begin{tikzpicture}\node[draw, thick, shape border rotate=90, isosceles triangle, isosceles triangle apex angle=60, fill=violet!70!white, fill opacity=1.0, node distance=1cm,minimum height=1.5em] at (0,0) {};\end{tikzpicture}}\xspace}

\newcommand{\miniagent}{\resizebox{2mm}{!}{\begin{tikzpicture}\node[draw, thick, shape border rotate=90, isosceles triangle, isosceles triangle apex angle=60, fill=violet!70!white, fill opacity=1.0, node distance=1cm,minimum height=1.5em] at (0,0) {};\end{tikzpicture}}\xspace}

\newcommand{\cA}{\mathcal{A}}
\newcommand{\cB}{\mathcal{B}}
\newcommand{\cC}{\mathcal{C}}

\newcommand{\cE}{\mathcal{E}}
\newcommand{\cF}{\mathcal{F}}

\newcommand{\cK}{\mathcal{K}}

\newcommand{\cM}{\mathcal{M}}
\newcommand{\cO}{\mathcal{O}}
\newcommand{\cP}{\mathcal{P}}
\newcommand{\cR}{\mathcal{R}}
\newcommand{\cS}{\mathcal{S}}
\newcommand{\cT}{\mathcal{T}}
\newcommand{\cU}{\mathcal{U}}
\newcommand{\cV}{\mathcal{V}}
\newcommand{\cX}{\mathcal{X}}
\newcommand{\cW}{\mathcal{W}}

\newcommand{\EEcp}[3]{\mathbb{E}_{#3}\left[#1\;\middle\lvert\;#2\right]}

\newcommand{\real}{\mathbb{R}}

\newcommand{\w}{\mathbf{w}}

\definecolor{mygreen}{HTML}{569e34}

\newcommand{\boldpsi}{\boldsymbol{\mathbf{\psi}}}

\theoremstyle{plain}
\newtheorem{theorem}{Theorem}

\usepackage[switch, modulo]{lineno}

\title{Planning with a Learned Policy Basis\\ to Optimally Solve Complex Tasks}
\author{
    David Kuric\equalcontrib\textsuperscript{\rm 1},
     Guillermo Infante\equalcontrib\textsuperscript{\rm 2},
    Anders Jonsson\textsuperscript{\rm 2},
    Vicen\c{c} G\'omez\textsuperscript{\rm 2},
    Herke van Hoof\textsuperscript{\rm 1}
}
\affiliations{
    \textsuperscript{\rm 1}
    AMLab, University of Amsterdam, Amsterdam, Netherlands\\
    \textsuperscript{\rm 2}
    AI\&ML group, Universitat Pompeu Fabra, Barcelona, Spain \\
    \{guillermo.infante,anders.jonsson,vicen.gomez\}@upf.edu, \{d.kuric,h.c.vanhoof\}@uva.nl 
}

\usepackage{bibentry}

\usepackage{cuted}

\begin{document}
\newcommand{\david}[1]{\textcolor{blue}{#1}}
\newcommand{\guille}[1]{\textcolor{teal}{#1}}

\maketitle
\begin{abstract}
Conventional reinforcement learning (RL) methods can successfully solve a wide range of sequential decision problems. However, learning policies that can generalize predictably across multiple tasks in a setting with non-Markovian reward specifications is a challenging problem. We propose to use successor features to learn a policy basis so that each (sub)policy in it solves a well-defined subproblem. In a task described by a finite state automaton (FSA) that involves the same set of subproblems, the combination of these (sub)policies can then be used to generate an optimal solution without additional learning. In contrast to other methods that combine (sub)policies via planning, our method asymptotically attains global optimality, even in stochastic environments.

\end{abstract}

\section{Introduction}
Autonomous agents that interact with an environment usually face tasks  that comprise complex, entangled behaviors over long horizons. Conventional reinforcement learning (RL) methods have successfully addressed this.  However, in cases when the agent is meant to perform several tasks across similar environments, training a policy for every task separately can be time-consuming and requires a lot of data. In such cases, the agent can utilize a method that has built-in generalization capabilities. One such method relies on the assumption that reward functions of these tasks can be decomposed into a linear combination of successor features \cite{Barreto2017}. When a new task is presented, it is possible to combine previously learned policies and their successor features to solve a new task. 
While combining such policies is guaranteed to be an improvement over any previously learned policy, it may not necessarily be optimal. 
However, as shown by \citet{Alegre2022}, 
one can leverage recent advancements in multi-objective RL to learn a set of policies that constitutes a policy basis to retrieve an optimal policy for any linear combination of successor features. 

While traditional RL methods rely on Markovian reward functions, defining a task using such a function can be challenging and sometimes impossible~\cite{Whitehead1995}. In scenarios where expressing the reward function in Markovian terms is not feasible, there has been a growing interest in alternative methods for task specification in recent years~\citep{Icarte2018, Camacho2019}. 
In our work we focus on developing a method that utilizes generalization capabilities of successor features in a settings with non-Markovian reward functions.

Prior techniques for such settings have been proposed in contexts where a set of propositional symbols enables the definition of high-level tasks using logic~\citep{Vaezipoor2021, ToroIcarte2019}
or finite state automata (FSA)~\cite{Icarte2018}. Like in hierarchical RL,
they are often based on decomposing tasks into sub-tasks and solving each sub-task independently~\citep{Dietterich2000, Sutton1999}. 
However, combining optimal solutions for sub-tasks may potentially result in a suboptimal overall policy. 
This is referred to as recursive optimality~\cite{Dietterich2000} or myopic policy~\cite{Vaezipoor2021}. 

To alleviate this issue, one can consider methods that condition the policy or the value function on the specification of the whole task \citep{UVF} and such approaches were recently also proposed for tasks with non-Markovian reward functions \cite{Vaezipoor2021}. However, the methods that specify the whole task usually rely on a blackbox neural network for planning when determining which sub-goal to reach next. This makes it hard to interpret the plan to solve the task and although  they show promising results in practice, 
it is unclear whether and when these approaches will generalize to a new task.


Instead, our work aims to use task decomposition without sacrificing global optimality to achieve predictable generalization. The method we propose learns a set of \textit{local} policies in sub-tasks such that their combination forms a \textit{globally optimal} policy for a large collection of problems described with FSAs. A new policy that solves any new task can then be created, without additional learning, by planning on a given FSA task description. Our contributions are:
 \begin{itemize}
    \item We propose to use successor features to learn a policy basis that is suitable for planning in stochastic domains.
    \item We develop a planning framework that uses such policy bases for zero-shot generalization to complex temporal tasks described by an arbitrary FSA.
    \item We prove that if the policies in this basis are optimal, our framework produces a globally optimal solution even in stochastic domains.
\end{itemize}

\section{Background and Notation} 
Given a finite set $\cX$, let $\Delta(\cX)=\{p\in\real^\cX:\sum_x p(x)=1,\linebreak p(x)\geq 0\;(\forall x)\}$ denote the probability simplex on $\cX$. Given a probability distribution $q\in\Delta(\cX)$, let $\mathrm{supp}(q)=\{x\in\cX:q(x) > 0\}\subseteq\cX$ denote the support of $q$. We abuse notation and let $\Delta(d)$ to represent the (standard) simplex in $\real^d$.

\subsection{Reinforcement Learning}
Reinforcement learning problems commonly assume an underlying Markov Decision Process (MDP). We define an MDP as the tuple $\cM = \langle\cS,\cE,\cA,\cR,\mathbb{P}_0, \mathbb{P},\gamma\rangle$ where $\cS$ is the set of states, $\cE$ is the set of exit states, $\cA$ is the action space, $\cR:\cS\times\cA\times\cS\rightarrow\real$ is the reward function, $\mathbb{P}_0 \in \Delta(\cS)$ is the probability distribution of initial states, $\mathbb{P}:\cS\times\cA\to\Delta(\cS)$ is the transition probability function and $0\leq\gamma<1$ is the discount factor. The set of exit states $\cE$ induces a set of terminal transitions $T = (\cS\setminus\cE)\times\cA\times\cE$.

The learning agent interacts in an episodic manner with the environment following a policy $\pi:\cS\rightarrow\Delta(\cA)$. At each timestep, the agent observes a state $s_t$, chooses the action $a_t\sim\pi(s_t)$, transitions to a new state $s_{t+1}\sim\mathbb{P}(\cdot\lvert s_t, a_t)$ and receives a reward $\cR(s_t, a_t, s_{t+1})$. The episode ends when the agent observes a terminal transition $(s_t, a_t, s_{t+1})\in T$ and a new episode starts with initial state $s_0\sim\mathbb{P}_0(\cdot)$.

The goal of the agent is to find an optimal policy $\pi^*$ that maximizes the expected discounted return, for any state-action pair ${(s,a)\in\cS\times\cA}$,
\begin{equation}
Q^\pi(s, a) = \EEcp{\sum_{i=t}^\infty \gamma^{i-t}\cR_i}{S_t=s, A_t=a}{\pi},
\label{eq:qfunction} 
\end{equation}where $\cR_i = \cR(S_i, A_i, S_{i+1})$. Hence, an optimal policy is
\begin{equation*}
  \pi^*\in \argmax_\pi Q^\pi(s, a)\;\;\forall(s,a)\in\cS\times\cA 
\end{equation*} with ties broken arbitrarily.
The action value function defined in Equation~\eqref{eq:qfunction} satisfies the recursive Bellman equation
\begin{equation}
     Q^\pi(s,a) = \mathbb{E}_{s'\sim\mathbb{P}(\cdot \lvert s, a)}\Big[\cR(s, a, s')+\gamma V^\pi(s')\Big],
\end{equation}
for any $(s,a)\in\cS\times\cA$. The state value function is obtained by averaging the action value function over the actions, $V^\pi(s)~=~\mathbb{E}_{a\sim\pi(s)}\left[ Q^\pi(s, a)\right]\;\forall s \in\cS$.
Throughout the paper we use $Q^*$ and $V^*$ to refer to the optimal, respectively, action and state value functions.
 
\subsection{Successor Features}

Successor features (SFs)~\citep{Dayan1993, Barreto2017} is a widely used RL representation framework that assumes the reward function is linearly expressible with respect to a feature vector,
\begin{equation}
  \cR^\w(s, a, s') = \w^\intercal\boldsymbol\phi(s, a , s').
  \label{eq:reward_sf}
\end{equation}

Here, $\boldsymbol\phi:\cS\times\cA\times\cS\rightarrow\real^{d}$ maps transitions to feature vectors and $\w\in\real^d$ is a weight vector. Every weight vector~$\w$ induces a different reward function and, thus, a task. The SF vector of a state-action pair $(s,a)\in\cS\times\cA$ under a policy $\pi$ is the expected discounted sum of future feature vectors: 
\begin{equation}
  \boldpsi^\pi(s, a) = \EEcp{\sum_{i=t}^\infty \gamma^{i-t} \boldsymbol\phi_i}{S_t = s, A_t = a}{\pi},
  \label{eq:sf}
\end{equation}
where $\boldsymbol\phi_i = \boldsymbol\phi(S_{i}, A_{i}, S_{i+1})$. The action value function for a state-action pair $(s, a)$ under policy $\pi$ can be efficiently represented using the SF vector. Due to the linearity of the reward function, the weight vector can be decoupled from the Bellman recursion. Following the definition of Equations~\eqref{eq:qfunction}~ and \eqref{eq:reward_sf}, the action value function in the SF framework can be rewritten as
\begin{align}
  Q^\pi_\w(s, a) &= \EEcp{\sum_{i=t} ^\infty \gamma^{i-t} \w^\intercal\boldsymbol\phi_i}{S_t = s, A_t = a}{\pi} \nonumber \\
                 & = \w^\intercal\EEcp{\sum_{i=t} ^\infty \gamma^{i-t} \boldsymbol\phi_i}{S_t = s, A_t = a}{\pi} \nonumber \\
                 &=  \w^\intercal \boldpsi^\pi(s, a) .
\label{eq:qfunction_sf}
\end{align}

The SF representation leads to \textit{generalized policy evaluation} (GPE) over multiple tasks~\cite{Barreto2020a}, and similarly, to \textit{generalized policy improvement} (GPI) to obtain new better policies~\cite{Barreto2017}.

A family of MDPs is defined as the set of MDPs that share all the components, except the reward function. This set is formally defined as 
\begin{equation*}
    \cM^{\boldsymbol{\phi}}\equiv\{\langle\cS,\cE,\cA,\cR_\w,\mathbb{P}_0, \mathbb{P},\gamma\rangle \lvert \cR_\w = \w^\intercal \boldsymbol{\phi}, \forall\w\in\real^d\}.
\end{equation*}

Transfer learning on families of MDPs is possible thanks to GPI. Given a set of policies $\Pi$, learned on the same family~$\cM^{\boldsymbol{\phi}}$, for which their respective SF representations have been computed, and a new task $\w'\in\real^d$, a GPI policy $\pi_{\text{GPI}}$ for any $s\in\cS$ is derived as 
\begin{equation}
    \pi_{\text{GPI}}(s) \in \argmax_{a\in\cA} \max_{\pi\in\Pi} Q^\pi_{\w'}(s, a).
    \label{eq:gpi}
\end{equation}

However, there is no guarantee of optimality for $\w'$.
A fundamental question to solve the so-called \textit{optimal policy transfer learning problem} is which policies should be included in the set of policies $\Pi$ so an optimal policy for any weight vector $\w\in\real^d$ can be obtained with GPI. 

\subsection*{Convex Coverage Set of Policies}

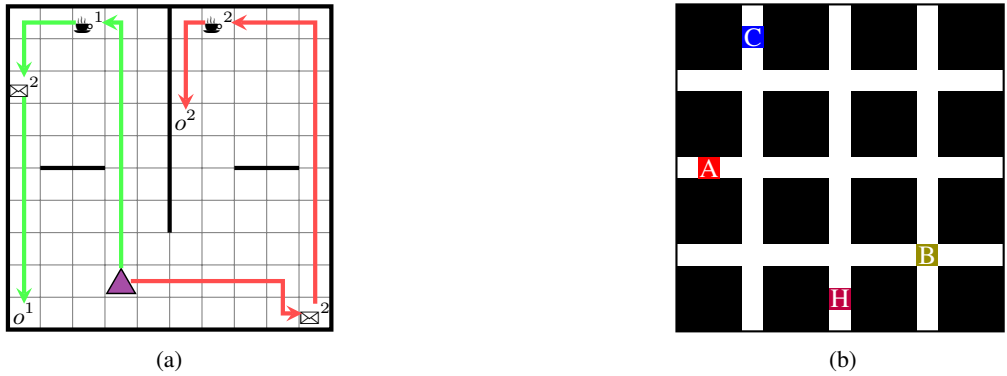
\begin{figure*}[!tb]
  \begin{subfigure}[t]{0.5\textwidth}
    \centering
    \begin{tikzpicture}[scale=0.43]
    \draw[step=1cm,gray] (0,0) grid (10, 10);
    \draw[ultra thick, fill=black] (5,10) -- (5, 3);
    \draw[ultra thick, fill=black] (1,5) -- (3, 5);
    \draw[ultra thick, fill=black] (7,5) -- (9, 5);

    \node at (0.5,0.5) {\small $o^1$};
    \node at (0.5,7.5) {\small $\text{\mail}^2$};
    \node at (2.5,9.5) {\small $\text{\coffee}^1$};
    \node at (6.5,9.5) {\small $\text{\coffee}^2$};
    \node at (5.5,6.5) {\small $o^2$};
    \node at (9.5,0.5) {\small $\text{\mail}^2$};

    \node at (3.5,1.5) {\small \agent};
    \draw[ultra thick, ->, >=stealth, draw=green!70!white] (3.5,1.9) -- (3.5,9.5) -- (2.9,9.5);
    \draw[ultra thick, ->, >=stealth, draw=green!70!white] (2.1,9.5) -- (0.5,9.5) -- (0.5,7.8) ;
    \draw[ultra thick, ->, >=stealth, draw=green!70!white] (0.5,7.2) -- (0.5,0.8);
     \draw[ultra thick, ->, >=stealth, draw=red!70!white] (3.8,1.5) -- (8.5, 1.5) -- (8.5, 0.5) -- (9.1, 0.5);
     \draw[ultra thick, ->, >=stealth, draw=red!70!white] (9.5,0.8) -- (9.5, 9.5) -- (6.9, 9.5);
    \draw[ultra thick, ->, >=stealth, draw=red!70!white] (6.1, 9.5) -- (5.5, 9.5) -- (5.5, 6.8);

    \draw[ultra thick] (0,0) rectangle (10,10);

\end{tikzpicture}
    \subcaption{}
    \label{fig:office_domain}
  \end{subfigure} 
  \hfill
  \begin{subfigure}[t]{0.5\textwidth}
    \centering
    \begin{tikzpicture}[scale=0.29]
    \fill[black] (0,0) rectangle ++ (3,3); 
    \fill[black] (4,0) rectangle ++ (3,3); 
    \fill[black] (8,0) rectangle ++ (3,3); 
    \fill[black] (12,0) rectangle ++ (3,3); 

    \fill[black] (0,4) rectangle ++ (3,3); 
    \fill[black] (4,4) rectangle ++ (3,3); 
    \fill[black] (8,4) rectangle ++ (3,3); 
    \fill[black] (12,4) rectangle ++ (3,3);

    \fill[black] (0,4) rectangle ++ (3,3); 
    \fill[black] (4,4) rectangle ++ (3,3); 
    \fill[black] (8,4) rectangle ++ (3,3); 
    \fill[black] (12,4) rectangle ++ (3,3);

    \fill[black] (0,8) rectangle ++ (3,3); 
    \fill[black] (4,8) rectangle ++ (3,3); 
    \fill[black] (8,8) rectangle ++ (3,3); 
    \fill[black] (12,8) rectangle ++ (3,3);

    \fill[black] (0,12) rectangle ++ (3,3); 
    \fill[black] (4,12) rectangle ++ (3,3); 
    \fill[black] (8,12) rectangle ++ (3,3); 
    \fill[black] (12,12) rectangle ++ (3,3);

    \fill[red] (1,7) rectangle ++ (1,1);
    \node at (1.5,7.5) {\color{white} A};
    \fill[blue] (3,13) rectangle ++ (1,1);
    \node at (3.5,13.5) {\color{white} C};
    \fill[olive] (11,3) rectangle ++ (1,1);
    \node at (11.5,3.5) {\color{white} B};
    \fill[purple] (7,1) rectangle ++ (1,1);
    \node at (7.5,1.5) {\color{white} H};



    \draw[ thick] (0,0) rectangle (15,15);

\end{tikzpicture}
    \subcaption{}
    \label{fig:delivery_domain}
  \end{subfigure}
  \caption{Depiction of the Office (a) and Delivery (b) environments, FSA task specification of the composite task in the Office domain and the FSA task specificiation of the sequential task in the Delivery domain (b). In (a) $\cP=\{\text{\coffee}, \text{\mail}, o\}$ and $\cE=\{\text{\coffee}^1,\text{\coffee}^2, \text{\mail}^1,\text{\mail}^2, o^1, o^2\}$. In (b), $\cE=\cP=\{A, B, C, H\}$.}
 \label{fig:domains}
\end{figure*}

The recent work of~\citet{Alegre2022} solves the optimal policy transfer learning problem. They draw the connection between the SF transfer learning problem and multi-objective RL (MORL). The pivotal fact is that the SF representation in Equation~\eqref{eq:sf} can be interpreted as a multidimensional value function and the construction of the aforementioned set of policies $\Pi$ can be cast as a multi-objective optimization problem.
 
 Consequently, the optimistic linear support (OLS) algorithm is extended with successor features in order to learn a set of policies that constitutes a \textit{convex coverage set} (CCS)~\cite{Roijers2015}. Their main result is the SFOLS algorithm (see Supplementary Material\footnote{Supplementary Material at {\texttt{\url{https://arxiv.org/abs/2403.15301}}}} for a full, technical description) in which a set $\Pi_\text{CCS}$ is built incrementally by adding (new) policies to such a set, until convergence. The set $\Pi_\text{CCS}$ contains all non-dominated policies in terms of their multi-objective value functions, where the dominance relation is defined over scalarized values ${V^\pi_\w = \mathbb{E}_{S_0\sim\mathbb{P}_0}\left[V^\pi_\w(S_0)\right]}$, and is characterized as
\begin{align}
  \Pi_\text{CCS} &= \{\pi\;\lvert\;\exists\w\;\text{s.t.}\;\forall {\boldpsi}^{\pi'},\; \w^\intercal {\boldpsi}^\pi {\;\geq\;} \w^\intercal{\boldpsi}^{\pi'} \} \nonumber\\
  &= \{ \pi \;\lvert\;\exists\w\;\text{s.t.}\;\forall {\pi'},\; {V^\pi_\w}{\;\geq\;} {V^{\pi'}_\w}\}.
\end{align} 
In every iteration $k$, SFOLS proposes a new weight vector $\w^k\in\Delta(d)$ for which an optimal policy (and its corresponding SF representation) is learned and added to $\Pi_\text{CCS}$ since it is sufficient to consider weights in $\Delta(d)$ to learn the full $\Pi_\text{CCS}$. The output of SFOLS is both $\Pi_\text{CCS}$ and the SF representation $\boldpsi^\pi$ for every $\pi\in\Pi_\text{CCS}$.
 
Intuitively, all policies in $\Pi_\text{CCS}$ are optimal in at least one task $\w \in \Delta(d)$.
The set $\Pi_\text{CCS}$ is combined with GPI, see Equation~\eqref{eq:gpi}, and upon convergence, for any (new) given task $\w'\in\real^d$, an optimal policy can be identified~\cite[cf. Theorem 2]{Alegre2022}.

\subsection{Propositional Logic}
We assume that environments are endowed with a set of high-level, boolean-valued propositional symbols $\cP$ and that they are associated with the set of exit states $\cE$ of a low-level MDP $\cM$. Every transition $\in\cS\times\cA\times\cS$ induces some propositional valuation (assignment of truth values) $2^\cP$. Such a valuation depends on the new state and occurs under a mapping ${\cO:\cS\rightarrow2^\cP}$ that is known to the agent. Nonetheless, only exit states $\varepsilon\in\cE$ make propositions true under $\cO$. We assume that that propositional symbols are mutually exclusive, and the agent cannot observe two symbols in the same transition. We say that a valuation $\Gamma$ satisfies a propositional symbol $p$, formally $\Gamma\vDash p$, if $p$ is true in $\Gamma$. 

\subsection{Finite State Automaton} 
\begin{figure}[!hbt]
  \centering
  \begin{subfigure}[h]{0.23\textwidth}
    \centering
    \begin{tikzpicture}[node distance=cm,on grid,every initial by arrow/.style={ultra thick,->, >=stealth}]
    \node[thick,state,initial above] (u_0) at (0,0) {$u_0$};
    \node[ thick,state]         (u_1) at (-0.7,-1.6)  {$u_1$};
    \node[ thick,state]         (u_2) at (0.7,-1.6)  {$u_2$};
    \node[circle,draw=black,minimum size=0.26cm,inner sep=0pt,fill=black] (t3) at (0,-3)  {};
    \node[text width=1cm ] at (-0.2,-2.95) {$u_T$};
    \path[thick,->, >=stealth] (u_0) edge node [left] {$\text{\coffee}$} (u_1);
    \path[thick,->, >=stealth] (u_0) edge node [right] {$\text{\mail}$} (u_2);
    \path[thick,->, >=stealth] (u_1) edge node [left] {$o$} (t3);
    \path[thick,->, >=stealth] (u_2) edge node [right] {$o$} (t3);
\end{tikzpicture}
    \caption{}
    \label{fig:office_fsa_disjunction}
  \end{subfigure}
  \hfill 
  \begin{subfigure}[h]{0.23\textwidth}
    \centering
    \begin{tikzpicture}[node distance=cm,on grid,every initial by arrow/.style={ultra thick,->, >=stealth}]
    \node[thick,state,initial above] (u_0) at (0,0) {$u_0$};
    \node[ thick,state]         (u_1) at (-0.7,-1.4)  {$u_1$};
    \node[ thick,state]         (u_2) at (0.7,-1.4)  {$u_2$};
    \node[ thick,state]         (u_3) at (-0.7,-2.8)  {$u_3$};

    \node[circle,draw=black,minimum size=0.26cm,inner sep=0pt,fill=black] (t3) at (0.9,-2.8)  {};
    \node[text width=1cm ] at (1.2,-2.3) {$u_T$};

    \path[thick,->, >=stealth] (u_0) edge node [left] {$\text{\coffee}$} (u_1);
    \path[thick,->, >=stealth] (u_0) edge node [right] {$\text{\mail}$} (u_2);
    \path[thick,->, >=stealth] (u_1) edge node [left] {$\text{\mail}$} (u_3);
    \path[thick,->, >=stealth] (u_2) edge node [right] {$\text{\coffee}$} (u_3);
    \path[thick,->, >=stealth] (u_3) edge node [below] {$\text{o}$} (t3);
    
\end{tikzpicture}
    \caption{}
    \label{fig:office_fsa_composite}
  \end{subfigure}
  \caption{Disjunction (a) and composite (b) FSA task specifications for the Office domain.}
 \label{fig:sample_fsas}
\end{figure}
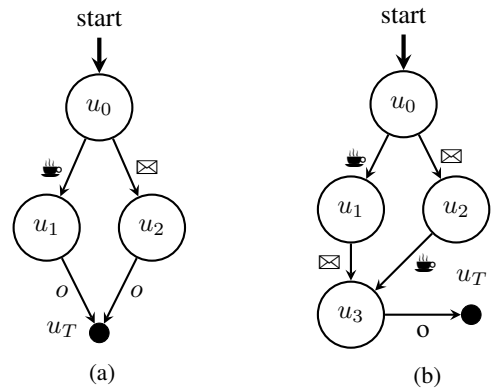

Task instructions can be specified via a finite state automaton. These are tuples ${\cF=\langle \cU,u_0,\cT,L,\delta\rangle}$ where $\cU$ is the finite set of states, $u_0\in\cU$ is the initial state, $\cT$ is the set of terminal states with $\cU\cap\cT=\emptyset$, $L:\cU\times(\cU\cup\cT)\rightarrow 2^\cP$ is a labeling function that maps FSA states transitions to truth values for the propositions and $\delta:\cU\rightarrow \{0, 1\}$ is a high-level reward function. Each transition among FSA states $(u, u')$ defines a subgoal. The agent has to observe some propositional valuation $L(u, u')$ in order to achieve it and FSA states can only be connected by a subgoal. E.g., in Figure~\ref{fig:office_fsa_disjunction}, the FSA state $u_0$ has two outgoing subgoals: getting mail (labeled as \mail) and getting coffee (labeled as \coffee). Non-existing transitions $(u, u')$ get mapped to $L(u, u')=\bot$. The reward function $\delta$ gives a reward larger than 0 only to terminal states. In other words, such a reward function is $\delta(u)=0\;\forall u\in\cU$ and $\delta(\mathbf{t})=1\;\forall \mathbf{t}\in\cT$. 

\section{Using Successor Features to Solve non-Markovian Reward Specifications}

We focus on the setting in which a low-level MDP is equipped with a reward structure like in Equation~\eqref{eq:reward_sf}. We let the low-level be represented by a family of MDPs $\cM^{\boldsymbol{\phi}}$, where each weight vector $\w\in\real^d$ specifies a low-level task. The agent receives high-level task specifications in the more flexible form of an FSA which permits the specification of non-Markovian reward structures. The combination of a low-level family of MDPs and a high-level FSA gives rise to a \textit{product MDP} $\cM'=\cF\times\cM^{\boldsymbol{\phi}}$ that satisfies the Markov property, and where the state space is augmented to be $\;\cU\times\cS$.

A product MDP $\cM'$ is a well-defined MDP. The agent now follows a policy $\mu:\cU\times\cS\rightarrow\Delta(\cA)$, that depends both on the FSA state and the underlying MDP state. $\cM'$ can be solved with conventional RL methods such as Q-learning~\cite{Watkins1992} by finding an optimal policy $\mu^*$ that maximizes
\begin{equation*}
Q^\mu(u, s, a) = \EEcp{\sum_{i=t}^\infty \gamma^{i-t}\cR_i}{U_t= u, S_t=s, A_t=a}{\mu}.
\end{equation*}
This is, however, impractical since policies should be retrained every time a new high-level task is specified. Exploiting the problem structure is essential for tractable learning, where components can be reused for new task specifications. The special reward structure of the low-level MDPs and our particular choice of feature vectors, later introduced, allow us to define an algorithm able to achieve a solution by simply planning in the space of augmented exit states $\;\cU\times\cE$. This inherently makes obtaining an optimal policy more efficient than solving the whole product MDP, as we reduce the number of states on which it is necessary to compute the value function.

 When presented with different task specifications (e.g.~Figure~\ref{fig:sample_fsas}), the agent may have to perform the same subtask at different moments of the plan or in different FSAs. We aim to provide agents with a collection of base behaviors that can be combined to retrieve the optimal behavior for the whole task.

In line with the previous reasoning, we introduce a two-step algorithm in which the agent first learns a $\Pi_{\text{CCS}}$ (a set of policies that constitute a CCS) on a well-specified representation of the environment. Then these (sub)policies are used to solve efficiently any FSA task specification on the propositional symbols of the environment. In what follows, we motivate the design of the feature vectors, explain our high-level dynamic programming algorithm and prove that it achieves the optimal solution.

 \subsubsection{Feature vectors} For a family of MDPs $\cM^{\boldsymbol{\phi}}$, feature vectors $\boldsymbol\phi(s, a, s')$ are \mbox{$\lvert\cE\rvert$-dimensional}. Each feature component $\boldsymbol\phi_j$ is associated with an exit state $\varepsilon_j\in\cE=\{\varepsilon_1,\ldots,\varepsilon_{\lvert\cE\rvert}\}$. Such vectors are built as follows. At terminal transitions $(s, a, \varepsilon_i)\in T$, $\boldsymbol\phi_{j} = 1$ when $j=i$ and $\boldsymbol\phi_{j}=0$ when $j\neq i$. For non-terminal transitions,  For non-terminal transitions, we just require that $\w^\intercal\boldsymbol\phi(s, a, s')<1$. In the case that ${\boldsymbol\phi(s, a, s')=\textbf{0}\in\real^{\lvert\cE\rvert}}$, the SF representation in Equation~\eqref{eq:sf} of each policy consists of a discounted distribution over the exit states. This indicates how likely it is to reach each exit state following such a policy. Furthermore, we require that $\cE\subset\text{supp}(\mathbb{P}_0)$ so the value functions at exit states are well-defined.

 \subsubsection{Example} In the office domain depicted in Figure~\ref{fig:office_domain}, the propositional symbols are $\cP=\{\text{\coffee}, \text{\mail}, \text{o}\}$ while the exit states $\cE=\{\text{\coffee}^1, \text{\coffee}^2,\text{\mail}^1,\text{\mail}^2,o^1,o^2\}$. Consequently, the same propositional symbol is satisfied at different exit locations, this is $\cO(\text{\coffee}^1)\vDash\text{\coffee}$ and $\cO(\text{\coffee}^2)\vDash\text{\coffee}$. In this case, $\boldsymbol\phi(s, a, s')\in\real^6$, is defined as the zero vector in $\real^6$ for every $s'\in\cS\setminus\cE$ and gets the corresponding vector component equal to $1$ when $s'\in\cE$. Figure~\ref{fig:sample_fsas} shows two different FSA task specifications for this domain, note that FSAs use symbols in $\cP$ to define the subgoals. The FSA in Figure~\ref{fig:office_fsa_disjunction} (disjunction) corresponds to `get coffee or mail, and then go to an office' and the one in Figure~\ref{fig:office_fsa_composite} (composite) to `get coffee and mail in any order, then go to an office'. 

\begin{algorithm}[!tb]
  \caption{SF-FSA-VI}
  \textbf{Input:} Low-level MDP $\cM^{\boldsymbol{\phi}}$, task specification $\cF$
  \begin{algorithmic}[1]
    \State Obtain $\Pi_\text{CCS}$ on $\cM^{\boldsymbol{\phi}}$.
    \State Initially  $\w^0(u) = \mathbf{0} \in\real^{\lvert\cE\rvert}\;\;\forall u\in\cU$.
   
    \While{not done}
      \For{$u \in \cU$}
        
        \State Update each $\w^{k+1}_j{(u)}$ with Equation~\eqref{eq:update_rule}.
       
      \EndFor
    \EndWhile
    
    \State \Return $\{\w^*(u)\;\forall u\in\cU\}$
  \end{algorithmic}
  \label{alg:online}
\end{algorithm}

\subsubsection{Algorithm} 

The solution to an FSA task specification implies solving a product MDP $\cM'=\cF\times\cM^{\boldsymbol{\phi}}$. Since we have the CCS, the optimal Q-function can be represented by a weight vector $\w^*$:
\begin{equation}
    Q^*_\w(u,s,a) = \smashoperator{\max_{\pi\in\Pi_\text{CCS}}} \w^*(u)^{\intercal}\boldsymbol{\psi}^\pi(s,a).
    \label{eq:extended_qfunction}
\end{equation}
for all $(u,s,a)\in\cU\times\cS\times\cA$. Here, $\w_j^*(u)$ indicates the optimal value from exit state $\varepsilon_j\in\cE$ for FSA state $u$. Then an optimal policy is defined as
\begin{equation}
    \mu^*_\w(u, s) \in \argmax_{a\in\cA} Q^*_\w(u,s,a)\;\forall(s,u)\in\cU\times\cS.
    \label{eq:hl-policy}    
\end{equation}
Therefore, we observe that finding the optimal weight vectors $\w^*(u)$, ${\forall u\in\cU}$ is enough for retrieving the optimal action value function of the product MDP $\cM'$ and, thus, an optimal policy.
 We can obtain this vector using a dynamic-programming approach similar to value iteration: 
\begin{align}
\w_j^{k+1}(u) =&  \max_a Q^*_\w\bigl(\tau(u,\cO(\varepsilon_j)),a\bigr) \\
              =& 
    \max_{a,\pi} \w^k\bigl(\tau(u,\cO(\varepsilon_j))\bigr)^{\intercal} \boldsymbol{\psi}^\pi (\varepsilon_j,a),  
    \label{eq:update_rule}
\end{align}
where $\tau(u,\cO(\varepsilon))\in\cU$ is the FSA state that results from achieving the valuation $\cO(\varepsilon)$ in $u$. We know that, $\w^k_j(u)=1$ if $\tau(u,\cO(j))=\textbf{t}$, per definition, since the high-level reward function $\delta(\mathbf{t})=1$. 
As a result, we propose SF-FSA-VI (see Algorithm~\ref{alg:online}) to extract an optimal policy for a product MDP. As $k\rightarrow\infty$, SF-FSA-VI converges to the optimal set of weight vectors $\{\w^*(u)\}_{u\in\cU}$ and, hence, to the optimal value function in Equation~\eqref{eq:extended_qfunction}.

\subsubsection{Proof of optimality} We first restate the following theorem from~\citet{Alegre2022}.

\begin{theorem}[Alegre, Bazzan, and Silva, 2022]
Let $\Pi$ be a set of policies such that the set of their expected SFs, $\Psi=\{\boldsymbol{\psi}^\pi\}_{\pi\in\Pi}$, constitutes a CCS. Then, given any weight vector $\w\in\real^d$, the GPI policy $\pi_\w^{GPI}(s) \in \arg \max_{a\in A} \max_{\pi\in\Pi} Q_\w^\pi(s,a)$ is
optimal with respect to ${\w: V_\w^{GPI} = V_\w^*}$.
\end{theorem}

\noindent
Applied to our setting, once the set of policies $\Pi_\text{CCS}$ and associated SFs have been computed, we can define an arbitrary vector $\w$ of rewards on the exit states, and use the CCS to obtain an optimal policy $\mu_\w^*$ and an optimal value function $V_\w^*$ without learning. We can then use composition by setting the reward of the exit states equal to the optimal value.

We aim to show that for each augmented state ${(u,s)\in\cU\times\cS}$, the value function output by our algorithm equals the optimal value of $(u,s)$ in the product MDP $\cM'=\cF\times\cM^{\boldsymbol{\phi}}$, i.e.~that $V_{\w(u)}(s)=V_{\cM'}^*(u,s)$. To do so, it is sufficient to show that the weight vectors $\{\w(u)\}_{u\in\cU}$ are optimal.

 Each element of $\w(u)$ is recursively defined as $\w_j(u)=V_{\w(\tau(u,\cO(\varepsilon_j)))}(\varepsilon_j)$. If all weight vectors are optimal, it holds that $V_{\w(\tau(u,\cO(\varepsilon_j)))}(\varepsilon_j)=V_{\cM'}^*(\w(\tau(u,\cO(\varepsilon_j))),\varepsilon_j)$ for each such exit state. Due to the above theorem, the value function $V_{\w(u)}$ is optimal for $\w(u)$. Due to composition that follows GPE and GPI, this means that the value of each internal state $s$ is optimal, i.e.~that $V_{\w(u)}(s)=V_{\cM'}^*(u,s)$.

It remains to show that the weight vectors $\{\w(u)\}_{u\in\cU}$ returned by the algorithm are indeed optimal. To do so it is sufficient to focus on the set of augmented exit states $\cU\times\cE$. We can state a set of optimality equations on the weight vectors as follows:
\begin{align*}
\w_j^*(u) &= V_{\w^*(\tau(u,\cO(\varepsilon)))}(\varepsilon_j)= \max_a Q^*(\tau(u,\cO(\varepsilon)),\varepsilon_j,a)\\
 &= \max_a\max_\pi {\boldsymbol{\psi}}^\pi(\varepsilon_j,a)^\intercal \w^*(\tau(u,\cO(\varepsilon))),
\end{align*}
where ${\boldsymbol{\psi}}^\pi(\varepsilon_j,a)=\sum_{s'}\mathbb{P}(s'|\varepsilon_j,a)\boldsymbol{\psi}^\pi(\varepsilon_j,a,s')$. Our termination condition implies that all subtasks take at least one time step to complete, and due to the discount factor $\gamma$, we have $\lVert\boldsymbol{\psi}(\varepsilon_j,a)\rVert_1<1$. Hence the update rule in Equation~\eqref{eq:update_rule} is a contraction and converges to the set of optimal weight vectors due to the Contraction Mapping Theorem.

\section{Experiments}
\begin{figure*}[htb]
 \begin{subfigure}[t]{0.5\textwidth}
    \centering
    \includegraphics[scale=0.28]{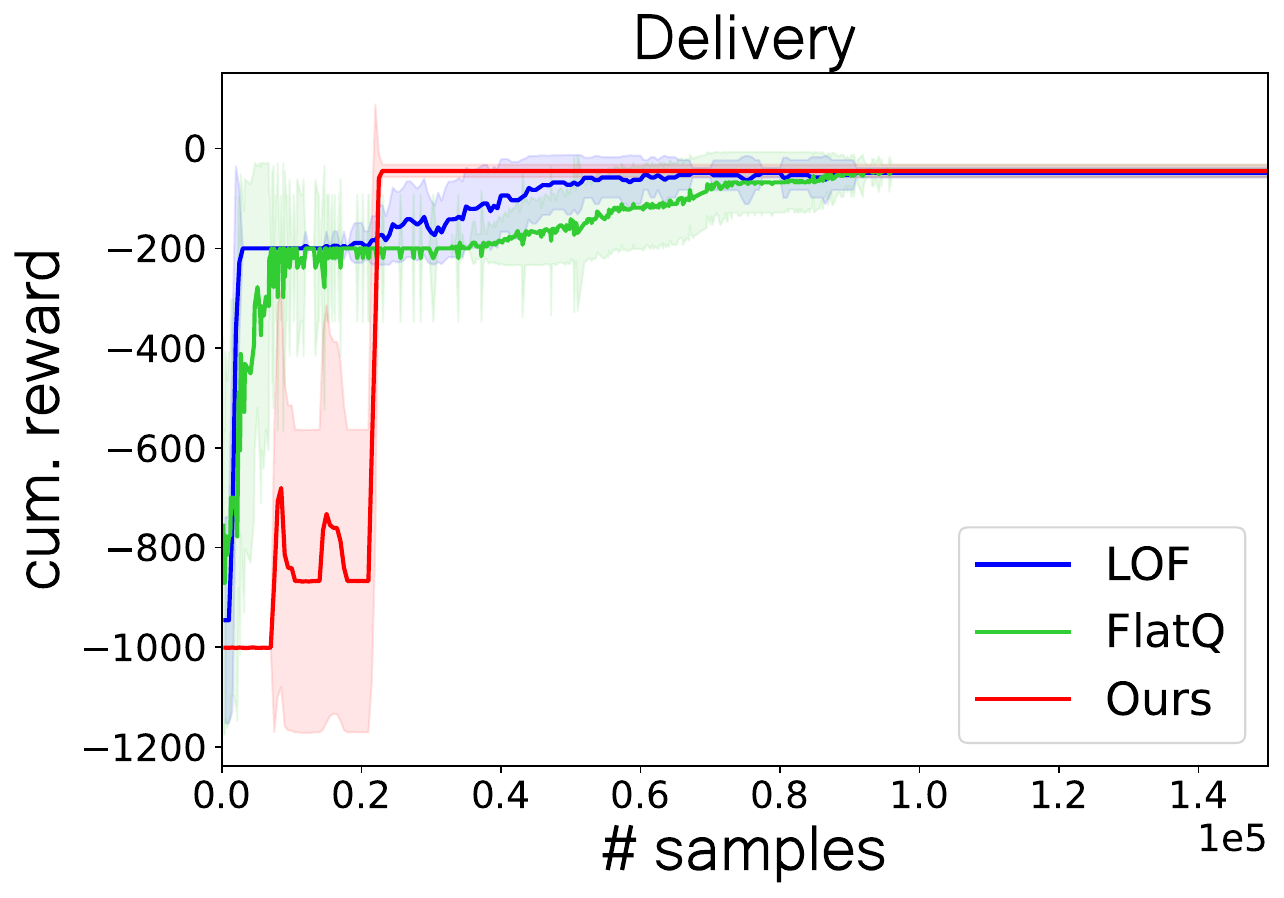}  
  \end{subfigure}
  \hfill
   \begin{subfigure}[t]{0.5\textwidth}
    \centering
    \includegraphics[scale=0.28]{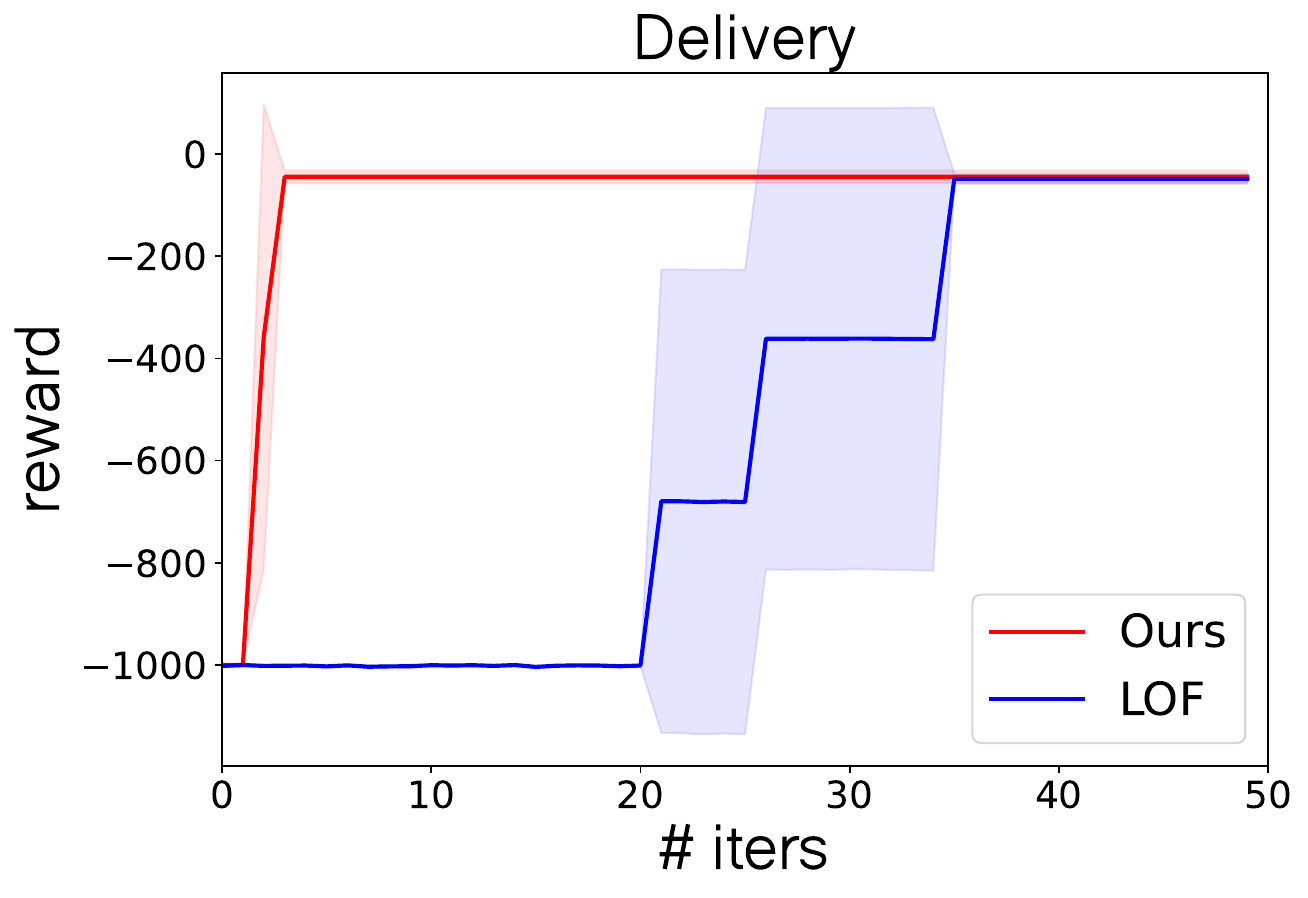}  
  \end{subfigure}
   \begin{subfigure}[b]{0.5\textwidth}
    \centering
    \includegraphics[scale=0.28]{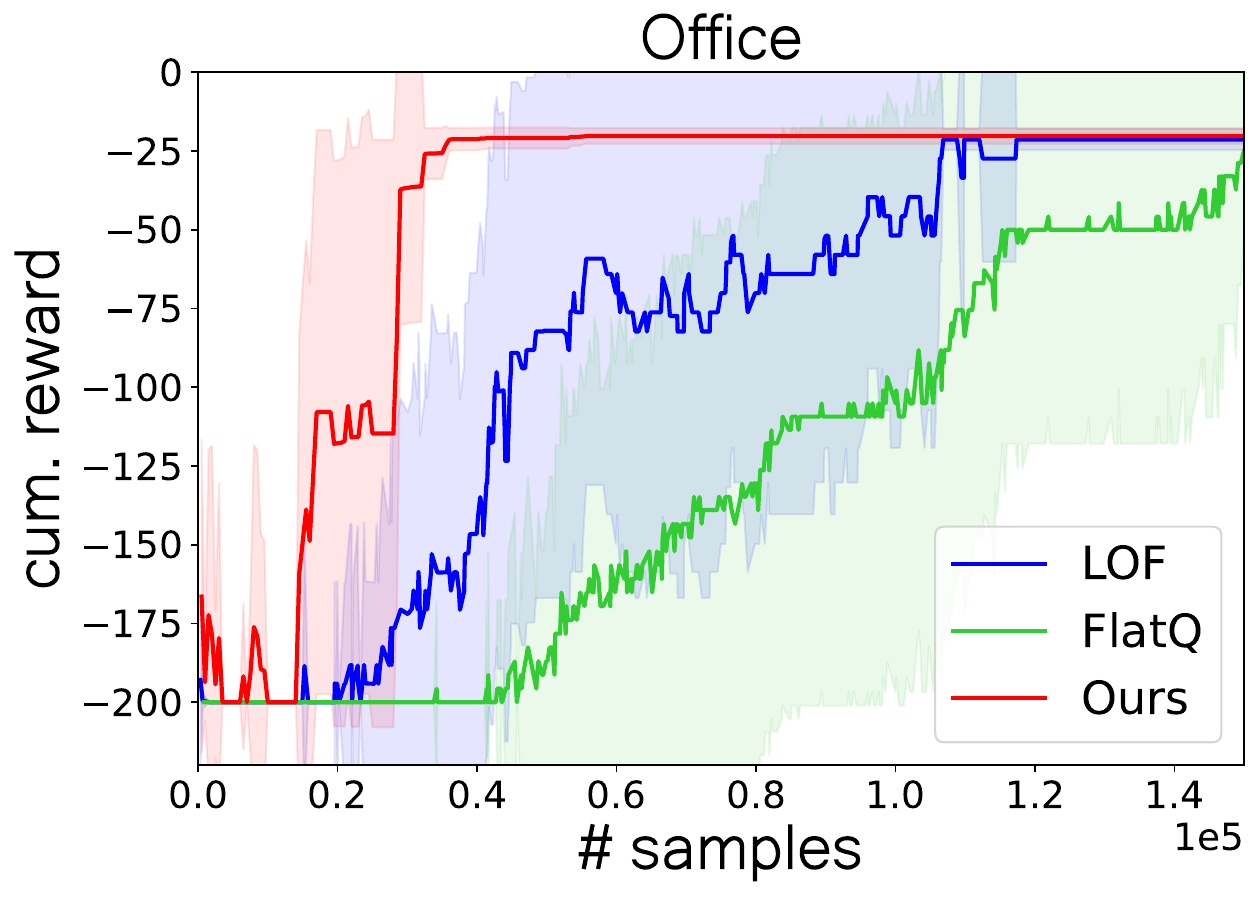}  
  \end{subfigure}
  \hfill
   \begin{subfigure}[b]{0.5\textwidth}
    \centering
    \includegraphics[scale=0.28]{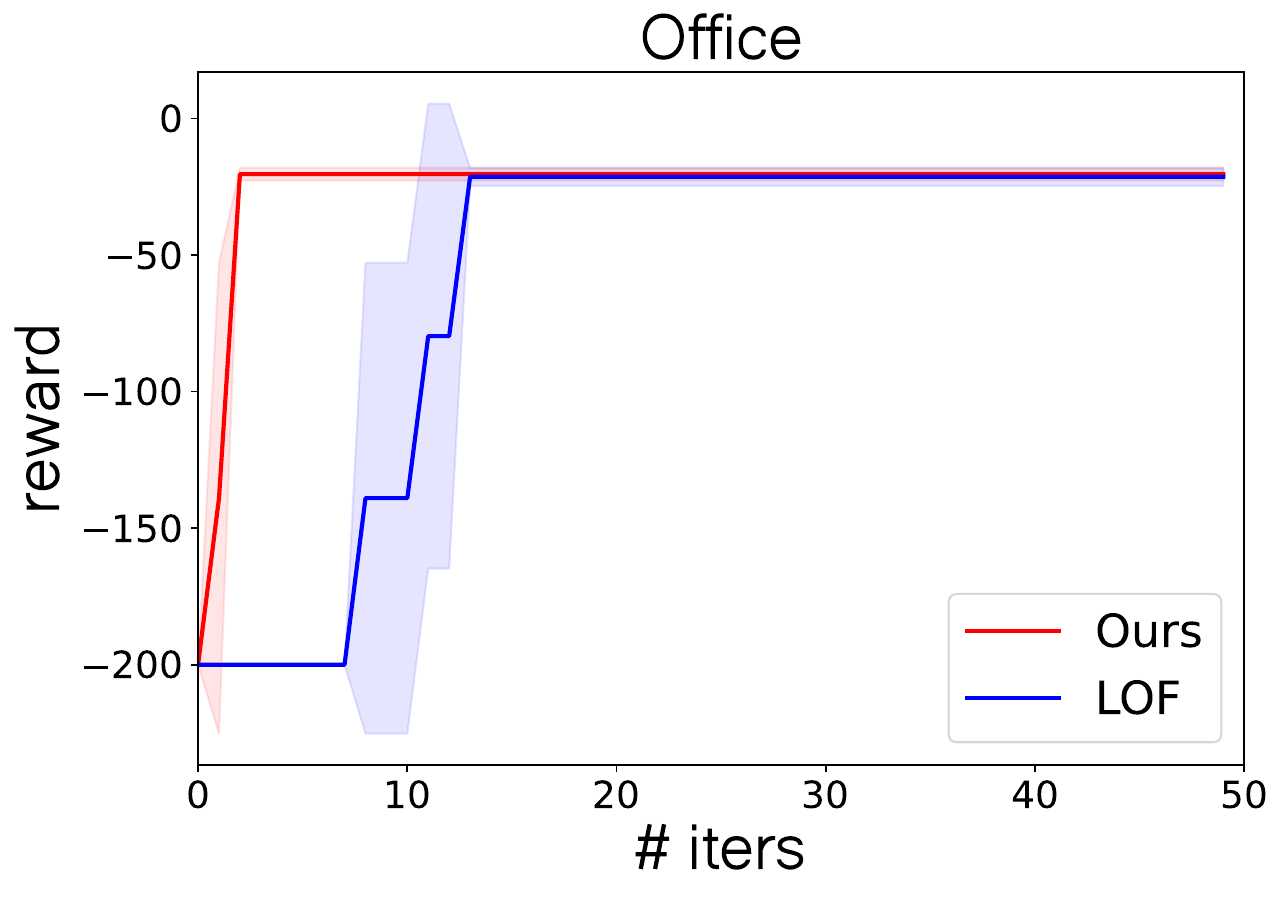}  
  \end{subfigure}
  \caption{Experimental results for learning (Delivery, top-left and Office, bottom-left) and compositionality (Delivery, top-right and Office, bottom-right). Results show the average performance and standard deviation over the three tasks and 5 seeds per task.}
  \label{fig:exp_results}
\end{figure*}
We test SF-FSA-VI in two complex discrete environments. At test time, we change the reward to $-1$ for every timestep and use the cumulative reward as the performance metric. We report two types of results. First, we are interested in observing the performance of the derived optimal policy, in Equation~\eqref{eq:hl-policy}, during the learning phase. For this, we fully retrain the high-level policy (lines 2-6 in Algorithm~\ref{alg:online}) every several interactions with the environment as $\Pi_\text{CCS}$ is being learned. Second, once the base behaviors are learned (this is once a complete $\Pi_\text{CCS}$ has been computed), we measure how many planning iterations SF-FSA-VI needs to converge to an optimal solution for different task specifications. In both cases, we compare against existing baselines.

\subsubsection{Environments and tasks} We use the Delivery domain~\cite{Araki2021} and a modified version of the Office domain~\cite{Icarte2018} as testbeds for our algorithm. Both environments are depicted in Figure~\ref{fig:domains} and present a propositional vocabulary that is rich enough to build complex tasks. In the Delivery domain there is a single low-level state associated with each of the propositional symbols, implying that ${\cE=\cP=\{A,B,C, H\}}$. The feature vectors are consistent with our design choice. For terminal transitions, they correspond to their one-hot encodings of the terminal states. There exist obstacle states (in black) for which, upon entering,  the feature vector is $\boldsymbol{\phi}(s,a,s')=\mathbf{-1000}\in\real^4$. This transforms in a large negative reward when multiplied with a corresponding weight vector $\w\in\real^{4}$. For regular grid cells (in white) $\boldsymbol{\phi}(s,a,s')=\mathbf{0}\in\real^{4}$. The Office domain is more complex since there are three propositional symbols $\cP=\{\text{\coffee}, \text{\mail}, o\}$ which can be satisfied at different locations, namely $\cE=\{\text{\coffee}^1, \text{\coffee}^2, \text{\mail}^1, \text{\mail}^1, o^1, o^2\}$. Here, there are no obstacle states and $\boldsymbol{\phi}(s,a,s')=\mathbf{0}\in\real^6$ for non-terminal transitions.

For each of the environments we define three different tasks: sequential, disjunction and composite (all described in the Supplementary Material). The sequential task is meant to show how our algorithm can indeed be effectively used to plan over long horizons, when the other two tasks show the ability of our method to optimally compose the base (sub)policies in complex settings. In natural language, the tasks in the Delivery domain correspond to: "go to $A$, then $B$, then $C$ and finally $H$"  (sequential), "go to $A$ or $B$, then $C$ and finally $H$" (disjunction) and "go to $A$ and $B$ in any order, then $B$, then $C$ and finally $H$" (composite). The agent has to complete the tasks by avoiding obstacles. The counterpart of these tasks in the Office environment are: "get a coffee, then pick up mail and then go to an office" (sequential), "get a coffee or mail, and then go to an office" (disjunction) and "get a coffee and mail in any order, and then go to an office" (composite). 
Our agent never learns how to solve these tasks, but rather learns the set of (sub)policies that constitutes the CCS. At test time, we provide the agent with the FSA task specification, extract a high-level optimal policy and test its performance on solving the task.

\subsubsection{Baselines} In the literature, we find the most similar approach to ours in the Logical Options Framework (LOF) ~\citep{Araki2021}. We thus use LOF and flat Q-learning on the product MDP as baselines. LOF trains one option per exit state, which are trained simultaneously using intra-option learning, and then uses a high-level value iteration algorithm to train a meta-policy that decides which option to execute in each of the MDP states. On the other hand, the latter learns the action value function in the flat product MDP, from which it extracts the policy. Under certain conditions, flat Q-learning converges to the optimal value function but, especially for longer tasks, it may take a large number of samples. Additionally, it is trained for a specific task, so it is not able to generalize to other task specifications. For LOF, we followed the implementation details prescribed by the authors. 

\subsection{Results}
\subsubsection{Learning} Empirical results for learning are shown in Figure~\ref{fig:exp_results} (top-left and bottom-left). The plots reflect how the different methods (ours, LOF and flat Q-learning) perform at solving an FSA task specification during the learning phase. In the case of SF-FSA-VI and LOF, the learning phase corresponds to obtaining the low level (sub)policies for $\Pi_\text{CCS}$ and the options, while for . Results are averaged over the three tasks (sequential, disjunction and composite) previously described for each environment. Each data point in the plots represent the cumulative reward obtained by a fully retrained policy with the current status of $\Pi_\text{CCS}$ and options. In both environments, SF-FSA-VI is the first to reach optimal performance. There exist, however, some differences between LOF and SF-FSA-VI. LOF trains all options simultaneously with intra-option learning. This means that, every transition $(s_t, a_t, s_{t+1})$ is used to update all options' value functions and policies. The learning of a $\Pi_\text{CCS}$, on the other hand, is done sequentially. A fixed sample budget per (sub)policy is set prior to learning, which can be seen as a hyperparameter. We use a total of $8\cdot 10^3$ samples per (sub)policy in both environments. A experience replay buffer is used to speed up the learning of the policy basis $\Pi_\text{CCS}$. Both options and the SF representation of (sub)policies are learned using Q-learning. Due to the incremental nature, at the beginning of the learning process there might be not enough policies in the basis $\Pi_\text{CCS}$ to construct a feasible solution. This is clearly observed in the Delivery domain (Figure~\ref{fig:exp_results}, top left), where at the early stages of the interaction, SF-FSA-VI achieves very low cumulative reward due to failing at delivering a solution. It is not until when there are enough (sub)policies in the basis that Algorithm~\ref{alg:online} attains a policy that solves the problem, which eventually converges to an optimal policy. Similarly, LOF converges to an optimal policy albeit it takes slightly longer to learn. In the more complex Office environment, results follow the same pattern. However, this environment breaks one of the of LOF requirements for optimality: to have a single exit state associated with each propositional predicate. In this problem, for each predicate there exist two exit states that can satisfy them. This makes LOF prone to converge to suboptimal solutions when SF-FSA-VI attains optimality. This is the case for the composite task, where LOF is short-sighted and returns a longer path (in red, Figure~\ref{fig:office_domain}) in contrast to ours that retrieves the optimal solution (in green, Figure~\ref{fig:office_domain}). This means that SF-FSA-VI is more flexible in the task specification. In this environment, our algorithm also converges faster with a more obvious gap with respect to LOF. In any case, learning (sub)policies or options is faster than learning directly on the flat product MDP, as flat Q-learning takes the longest to converge.
\subsubsection{Planning} Figure~\ref{fig:exp_results} top-right and bottom-right show how fast SF-FSA-VI and LOF can plan for an optimal solution. Results are again averaged for the three tasks for each environment. Here, a complete policy basis $\Pi_\text{CCS}$ has been previously computed, as well as the option's optimal policies.  In LOF, the cost of each iteration of value iteration is $\lvert\cU\rvert\times\lvert\cS\rvert\times \lvert\cK\rvert$, where $\cK$ is the set of options, while for the Algorithm~\ref{alg:online} we propose it is $\lvert\cU\rvert\times\lvert\cE\rvert\times\lvert\Pi_\text{CCS}\rvert$. By definition, the number of options is equivalent to the number of exit states $\lvert\cK\rvert=\lvert\cE\rvert$, so a single iteration of SF-FSA-VI is more efficient than LOF whenever $\lvert\Pi_\text{CCS}\rvert \ll \lvert\cS\rvert$. In our experiments, the sizes of the CCS are $15$ and $12$ for the Delivery and Office domains, respectively, while the sizes of the state spaces are of $225$ and $121$. Therefore, since our algorithm needs fewer, shorter iterations during planning, it outperforms LOF in terms of planning speed in both domains when composing the global solution. This can be observed in the plots for both environments. 

\subsubsection{Policy basis over options} 
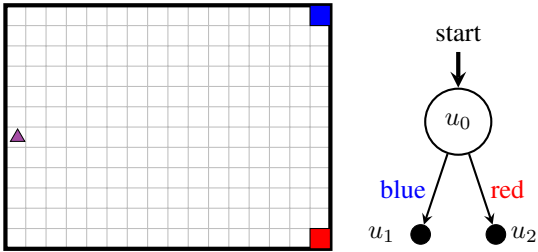
\begin{figure}[!tb]
  \centering
  \begin{subfigure}[t]{0.23\textwidth}
    \centering
    \begin{tikzpicture}[scale=0.54]

    \draw[step=0.5cm,lightgray] (0,0) grid (8, 6);
    \draw[ultra thick] (0,0) rectangle (8,6);

    \node at (0.3, 2.8) {\miniagent};
    \draw[fill=blue] (7.5,5.5) --  (7.5,6) -- (8,6) -- (8,5.5) -- cycle;
    \draw[fill=red]  (7.5,0) --  (7.5,0.5) -- (8,0.5) -- (8,0) -- cycle;


\end{tikzpicture}
  \end{subfigure}
  \hfill
  \begin{subfigure}[t]{0.23\textwidth}
    \centering
    \begin{tikzpicture}[node distance=cm,on grid,every initial by arrow/.style={ultra thick,->, >=stealth}]
    \node[thick,state,initial above] (u_0) at (0,0) {$u_0$};
    \node[circle,draw=black,minimum size=0.26cm,inner sep=0pt,fill=black] (t1) at (-0.5,-1.5)  {};
    \node[circle,draw=black,minimum size=0.26cm,inner sep=0pt,fill=black] (t2) at (0.5,-1.5)  {};
    \node[text width=1cm ] at (-0.7,-1.5) {$u_1$};
    \node[text width=1cm ] at (1.2,-1.5) {$u_2$};
    \path[thick,->, >=stealth] (u_0) edge node [left] {\color{blue} blue} (t1);
    \path[thick,->, >=stealth] (u_0) edge node [right] {\color{red} red} (t2);
\end{tikzpicture}
  \end{subfigure}
  \caption{Double Slit environment (left) and FSA task specification to reach either goal locations blue or red.}  
 \label{fig:double_slit}
\end{figure}

In deterministic environments, it is sufficient to learn the (sub)policies associated with the extrema weights (i.e. those (sub)policies that reach each of the exit states individually) to find a globally optimal policy via planning. In such cases, it may not be necessary to learn a full CCS. That is why, approaches that use the options framework such as LOF traditionally define one option per subgoal. However, there are scenarios, in which these approaches will not find optimal policy. This is the case for most stochastic environments. For example, consider the very simple domain of Double Slit in and the FSA task specification in Figure~\ref{fig:double_slit}. In this environment, there are two exit states $\cE=\{{\color{blue} \text{blue}}, {\color{red}\text{red}}\}$. The agent starts in the leftmost column and middle row. At every timestep, the agent chooses an action amongst $\{\text{UP}, \text{RIGHT}, \text{DOWN}\}$ and is pushed one column to the right in addition to moving in the chosen direction, except in the last column. If the agent chooses RIGHT, he moves an extra column to the right. At every timestep there is a random wind that can blow the agent away up to three positions in the vertical direction. The FSA task specification represents a task in which the agent is indifferent between achieving either of the goal states. Since the RIGHT action brings the agent closer to both goals, the optimal behavior in this case is to commit to either goal as late as possible. In this setting, methods that use one policy per sub-goal, such as LOF, train two policies to reach both goals. This means that the agent has to commit to one of the goals from the very beginning, which hurts the performance as it has to make up for the consequences of the random noise. On the other hand, the CCS used by SF-FSA-VI will contain an additional policy that is indifferent between two goals. This leads to a performance gap as our approach achieves an average accumulated reward of $-19.7\pm3.65$ and LOF $-22.70\pm 5.72$.

\section{Related Work}
One of the key distinctions in our research compared to prior studies is the optimality of the final solution. As noted by \citet{Dietterich2000}, hierarchical methods usually have the capability to achieve hierarchical, recursive, or global optimality. The challenge that often arises when sub-task policies are trained in isolation is that the combination of these locally optimal policies does not lead to a globally optimal policy but a recursively \cite{Dayan1992} or hierarchically optimal policy \cite{Sutton1999, mann2015approximate, Araki2021}.  To tackle this challenge, our approach relies on acquiring a set of low-level policies for each sub-task and employing planning to identify the optimal combination of low-level policies when solving a particular task. By learning the CCS with OLS \citep{roijers2014linear} in combination with high-level planning our approach ensures that globally optimal policy is found. In this regard, the work of \citet{Alegre2022} is of particular interest as it was the first work that used OLS and successor features \cite{Barreto2017} for optimal policy transfer learning. However, this method has only applied in a setting with Markovian reward function and has not been used with non-Markovian task specifications or high-level planning. 

On the other hand, many recent approaches proposed to use high-level task specifications in the form of LTL~\citep{Icarte2018b, kuo2020encoding, Vaezipoor2021, Jothimurugan2021}, or similar formal language specifications~\citep{ToroIcarte2019,Camacho2019, Araki2021, Icarte2022} to learn policies. However, the majority of the methods in this area are designed for single-task solutions, with only several focusing on acquiring a set of policies that is capable of addressing multiple tasks \cite{Icarte2018b, Leon2020, kuo2020encoding, Araki2021, Vaezipoor2021}. But, in contrast to our approach, they do not guarantee optimality of the solution.

From these works, our approach is the most similar to the Logical Options Framework \cite{Araki2021}. The main difference is that LOF trains a single policy for each sub-goal, resulting in a set of learned policies that is either smaller than or equal to the set acquired through SF-FSA-VI. While employing one policy per sub-goal proves sufficient for obtaining a globally optimal policy through planning in deterministic environments~\citep{Wen2020}, this may not hold true in stochastic environments, as our experiments demonstrate. In such instances, the policies generated by LOF are hierarchically optimal but fall short of global optimality.

Two notable examples from aforementioned works on multi-task learning with formal language specifications are the works of \citet{Icarte2018b} and \citet{Vaezipoor2021}. The former struggles with generalizing to unseen tasks, because it uses LTL progression to determine which sub-tasks need to be learned to solve given tasks. The Q-functions that are subsequently learned for each LTL sub-task will therefore not be useful for a new task if its sub-tasks were not part of the training set. Such limitation does not apply to the latter as it instead encodes the remaining LTL task specification using a neural network and conditions the policy on this LTL embedding. While this approach may be more adaptable to tasks with numerous propositions or sub-goals, it risks generating sub-optimal policies as it relies solely on the neural network to select the next proposition to achieve, without incorporating planning. Additionally, since the planning is implicitly done by the neural network, the policy is less interpretable than when explicit planning is used.

The method we propose can be viewed as a method for composing value functions through successor features, akin to previously proposed approaches for composition of value functions and policies ~\citep{Niekerk2019, Barreto2019, NangueTasse2020, Infante2022}. In the work of ~\citet{Infante2022}, which is the closest to our work, the authors propose to learn a basis of value functions that can be combined to form an optimal policy. However, unlike SF-FSA-VI, their approach only works in a restricted class of linearly-solvable MDPs. Lastly, since our approach uses the values of exit states for planning it is also related to planning with exit profiles \citep{Wen2020}. The CCS that we propose to use as a policy basis in our work can be seen as a collection of policies that are optimal for all possible exit profiles.

\section{Discussion and Conclusion}

In this work, we address the problem of finding optimal behavior for new non-Markovian goal specifications in known environments. To do so, we introduce a novel approach that uses successor features to learn a policy basis, that can subsequently be used to solve any unseen task specified by an FSA with the set of given predicates $\cP$ by planning. SF-FSA-VI is the first that can provably generalize to such new task specification without sacrificing optimality in both deterministic and stochastic environments.

The experiments show that SF-FSA-VI offers several advantages over previous methods. First, due to the use of SF, it allows for faster composition of the high-level value function since it drastically reduces the number of states to plan on. Secondly, thanks to using a CCS over a set of options SF-FSA-VI achieves optimality even in stochastic environments (as shown in the Double Slit example). Lastly, we do not require that there exists a single exit state per predicate which permits more flexible task specification while at the same time allowing deployment in more complex environments. 

A limitation of our approach could be the need to construct a full CCS if one wants to attain global optimality. While the construction of CCS is not timecomsuming for environments with several exit states presented in our work, the computation cost of finding the full CCS could become too large for environments with many exit states. In such case one could instead learn a partial CCS at the cost of a bounded decrease in performance \citep{Alegre2022} or consider splitting the environment into smaller parts with fewer exit states. While our experiments only considered discrete environments, SF-FSA-VI should also be applicable in continuous environments with minor adjustments. These include: using an contiguous set of states instead of a single exit state and using reward shaping to facilitate learning in sparse reward setting.

\section{Acknowledgements}

This publication is part of the action CNS2022-136178 financed by MCIN/AEI/10.13039/501100011033 and by the European Union Next Generation EU/PRTR.
This work has been co-funded by MCIN/AEI/10.13039/501100011033 under the Maria de Maeztu Units of Excellence Programme (CEX2021-001195-M).
Anders Jonsson is partially supported by the EU ICT-48 2020 project TAILOR (No. 952215), AGAUR SGR, and the Spanish grant PID2019-108141GB-I00. David Kuric acknowledges travel support from the ELISE Network funded by European Union’s Horizon 2020 research and innovation programme (GA No 951847).

\bibliography{aaai24}
\clearpage

\begin{strip}%
 \centering
 \LARGE {\bf Supplementary Material}\\
\end{strip}
\setcounter{secnumdepth}{1}

\section{Computation of a $\Pi_{\text{CCS}}$} 
The successor features (SF) extension  of the optimistic linear support (OLS) algorithm (SFOLS,~\citet{Alegre2022}) that is used in our work to compute $\Pi_{\text{CCS}}$ is fully described in Algorithm~\ref{alg:sfols}. 

It utilizes the value of the \textit{set max policy} (SMP,~ \citet{zahavy2021discovering}), which is a commonly used, weaker approach for transfer learning in multi-objective RL. For a given weight vector $\w$ and a set of policies $\Pi$, it is defined as:
\begin{align*}
    \pi^{\text{SMP}}_\w(s) = \pi'(s),\text{where } \pi'=\argmax_{\pi\in\Pi}V^\pi_\w
\end{align*}
and the value of this policy is $V^{\text{SMP}}_\w = \max_{\pi\in\Pi} V^\pi_\w$.

The algorithm constructs $\Pi_{\text{CCS}}$ incrementally. Starting with the extremum points of weights simplex $\cC^d$, it sequentially processes weights from its weight priority queue $Q$. In each iteration, an (optimal) policy and its successor feature representation are found for the selected weight $\w$. If the successor features of this policy are different from the successor features of policies currently in the $\Pi_{\text{CCS}}$, the new policy is added to the $\Pi_{\text{CCS}}$ and the weight priority queue $Q$ is adjusted. This adjustment has two steps. Firstly, the weights for which the new policy performs better than all current policies are removed and secondly, the new corner weights found with Algorithm~\ref{alg:cornerweights} are added to $Q$ with corresponding priority computed with Algorithm \ref{alg:estimateimprov}.

\begin{algorithm}[!htb]
  \caption{SFs Optimistic Linear Support (SFOLS)}
  \textbf{Initialize:} $\Pi_\text{CCS}\leftarrow \{\}, \Psi\leftarrow \{\}, \cW\leftarrow \{\}, Q\leftarrow \{\}$
  \begin{algorithmic}[1]
    \For{each extremum weight vector $\w_e\in\cC^d$}
        \State Add $\w_e$ to $Q$ with max. priority
    \EndFor

    \Repeat
    \State $\w\leftarrow$ pop weight with max. priority in $Q$
    \State $\pi, \boldsymbol{\psi}^\pi\leftarrow$ Solve $\langle\cS,\cE,\cA,\cR_\w,\mathbb{P}_0, \mathbb{P},\gamma\rangle$
    \State Add $\w$ to $\cW$
    \If{$\boldpsi^\pi\notin\Psi$}
        \State Remove from $Q$ all $\w'$ s.t. ${\w'}^\intercal\boldpsi^\pi > V^\text{SMP}_{\w'}$
        \State $X\leftarrow$ CornerWeights($\boldpsi^\pi,\w,\Psi$)
        \State Add $\Psi^\pi$ to $\Psi$ and $\pi$ to $\Pi_\text{CCS}$
        \For{$\w'\in X $}
            \State $\Delta(\w')\leftarrow$ EstimateImprovement($\w',\Psi $)
            \State Add $\w'$ to $Q$ with priority  $\Delta(\w')$
        \EndFor
    \EndIf
    \Until{$Q$ is empty}
    \State \Return $\Pi_{\text{CCS}}, \Psi$
  \end{algorithmic}
  \label{alg:sfols}
\end{algorithm}
The runtime complexity of the SFOLS algorithm depends heavily on the number of policies that have to be trained and considered for $\Pi_\text{CCS}$. It is  the same as the number of corner weights that must be analyzed in the original OLS for which \citet{Roijers2015} provide a following bound: 
\begin{equation*}
    O( \binom{\lvert CCS\rvert \lfloor\frac{\lvert\cE\rvert + 1}{2}\rfloor}{\lvert CCS\rvert - \lvert\cE\rvert} + \binom{\lvert CCS\rvert \lfloor\frac{\lvert\cE\rvert + 2}{2}\rfloor}{\lvert CCS\rvert - \lvert\cE\rvert}).
\end{equation*}
Scaling to many objectives can thus be prohibitive but should be possible by sacrificing optimality and using $\epsilon\text{-}CCS$ \citep{Alegre2022}.

\begin{algorithm}[!htb]
  \caption{EstimateImprovement}
  \textbf{Input:} New weight vector $\w$, $\Psi$, $\cW$
  \begin{algorithmic}[1]
    \State Let $\Bar{V}^*_{\w'}$ be the optimistic upper bound on $V^*_{\w'}$ computed by the following linear program 
    \begin{align*}
        \max\;&\w^\intercal\boldpsi\\
        \text{subject to}\;&\w'^\intercal\boldpsi\leq V^{\text{SMP}}_{\w'}\;\;\forall\;\w'\in\cW
    \end{align*}
    \State $\Delta(\w)\leftarrow \Bar{V}^*_{\w'} - V^{\text{SMP}}_{\w'}$
    \State \Return $\Delta(\w)$
  \end{algorithmic}
  \label{alg:estimateimprov}
\end{algorithm}

\begin{algorithm}[!bth]
  \caption{CornerWeights}
  \textbf{Input:} New SF vector $\boldpsi^\pi$, current weight vector $\w$, current set $\Psi$
  \begin{algorithmic}[1]
    \State Let $\cW_{\text{del}}$ be the set of obsolete weights removed from $Q$ in line 8 of Algorithm~\ref{alg:sfols}
    \State Add $\w$ to $\cW_{\text{del}}$
    \State $\cV_{\text{rel}}\leftarrow\{\boldpsi^\pi \lvert \boldpsi^\pi \in\argmax_{\boldpsi^\pi\in\Psi} \w'^\intercal\boldpsi^\pi\}$ for at least one $\w'\in\cW_{\text{del}}$
    \State $\cB_{\text{rel}}\leftarrow$ the set of boundaries of the weight simplex $\cC^d$ involved in any $\w'\in\cW_{\text{del}}$
    \State $X\leftarrow\{\}$
    \For{each subset $Y$ of $d-1$ elements from $\cV_{\text{rel}}\cup\cB_{\text{rel}}$}
    \State $\w_c\in\cC^d$ where $\boldpsi^\pi$ intersects with boundaries in $Y$
    \State Add $\w_c$ to $\cC_c$
    \EndFor

    \State\Return $X$

  \end{algorithmic}
  \label{alg:cornerweights}
\end{algorithm}

\section{Task specifications}
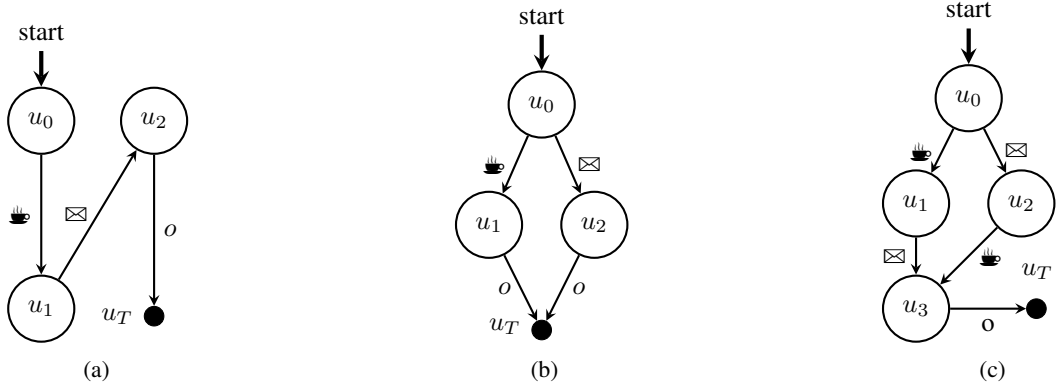
\begin{figure*}[!hbt]
  \begin{subfigure}[t]{0.33\textwidth}
    \centering
    \begin{tikzpicture}[node distance=cm,on grid,every initial by arrow/.style={ultra thick,->, >=stealth}]
    \node[thick,state,initial above] (u_0) at (0,0) {$u_0$};
    \node[thick,state]         (u_1) at (0,-2.5)  {$u_1$};
    \node[thick,state]         (u_2) at (1.5, 0)  {$u_2$};

    \node[circle,draw=black,minimum size=0.26cm,inner sep=0pt,fill=black] (t3) at (1.5,-2.6)  {};
    \node[text width=1cm] at (1.3,-2.6) {$u_T$};
    
    \path[thick,->, >=stealth] (u_0) edge node [left] {$\text{\coffee}$} (u_1);
    \path[thick,->, >=stealth] (u_1) edge node [left] {$\text{\mail}$} (u_2);
    \path[thick,->, >=stealth] (u_2) edge node [right] {$o$}(t3);
\end{tikzpicture}
    \subcaption{}
    \label{fig:officetasksequential}
  \end{subfigure}
  \hfill
  \begin{subfigure}[t]{0.33\textwidth}
    \centering
    \begin{tikzpicture}[node distance=cm,on grid,every initial by arrow/.style={ultra thick,->, >=stealth}]
    \node[thick,state,initial above] (u_0) at (0,0) {$u_0$};
    \node[ thick,state]         (u_1) at (-0.7,-1.6)  {$u_1$};
    \node[ thick,state]         (u_2) at (0.7,-1.6)  {$u_2$};
    \node[circle,draw=black,minimum size=0.26cm,inner sep=0pt,fill=black] (t3) at (0,-3)  {};
    \node[text width=1cm ] at (-0.2,-2.95) {$u_T$};
    \path[thick,->, >=stealth] (u_0) edge node [left] {$\text{\coffee}$} (u_1);
    \path[thick,->, >=stealth] (u_0) edge node [right] {$\text{\mail}$} (u_2);
    \path[thick,->, >=stealth] (u_1) edge node [left] {$o$} (t3);
    \path[thick,->, >=stealth] (u_2) edge node [right] {$o$} (t3);
\end{tikzpicture}
    \subcaption{}
    \label{fig:officetaskdisjunction}
  \end{subfigure}
  \hfill
  \begin{subfigure}[t]{0.33\textwidth}
    \centering
    \begin{tikzpicture}[node distance=cm,on grid,every initial by arrow/.style={ultra thick,->, >=stealth}]
    \node[thick,state,initial above] (u_0) at (0,0) {$u_0$};
    \node[ thick,state]         (u_1) at (-0.7,-1.4)  {$u_1$};
    \node[ thick,state]         (u_2) at (0.7,-1.4)  {$u_2$};
    \node[ thick,state]         (u_3) at (-0.7,-2.8)  {$u_3$};

    \node[circle,draw=black,minimum size=0.26cm,inner sep=0pt,fill=black] (t3) at (0.9,-2.8)  {};
    \node[text width=1cm ] at (1.2,-2.3) {$u_T$};

    \path[thick,->, >=stealth] (u_0) edge node [left] {$\text{\coffee}$} (u_1);
    \path[thick,->, >=stealth] (u_0) edge node [right] {$\text{\mail}$} (u_2);
    \path[thick,->, >=stealth] (u_1) edge node [left] {$\text{\mail}$} (u_3);
    \path[thick,->, >=stealth] (u_2) edge node [right] {$\text{\coffee}$} (u_3);
    \path[thick,->, >=stealth] (u_3) edge node [below] {$\text{o}$} (t3);
    
\end{tikzpicture}
    \subcaption{}
    \label{fig:officetaskcomposite}
  \end{subfigure}
  \caption{Finite state automatons for the Office domain (sequential (a), disjunction (b) and composite (c)) tasks.}
 \label{fig:fsas_office}
\end{figure*}

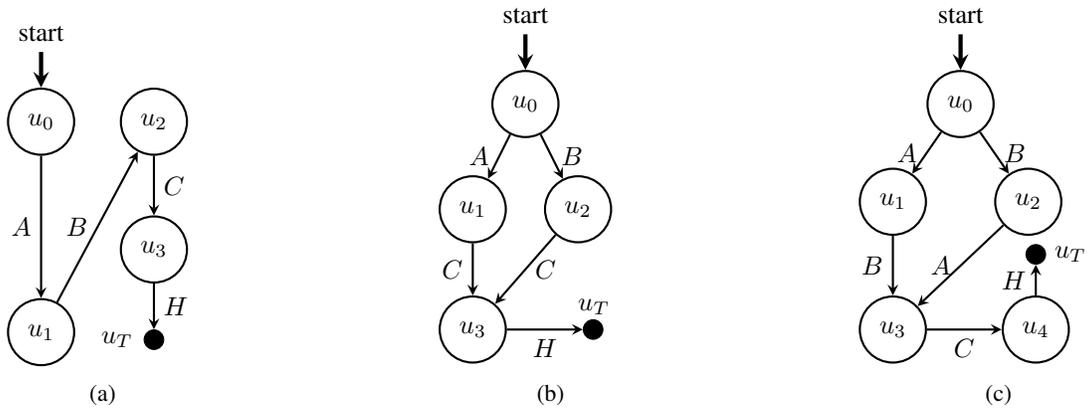
\begin{figure*}[!hbt]
  \begin{subfigure}[t]{0.33\textwidth}
    \centering
    \begin{tikzpicture}[node distance=cm,on grid,every initial by arrow/.style={ultra thick,->, >=stealth}]
    \node[thick,state,initial above] (u_0) at (0,0) {$u_0$};
    \node[thick,state]         (u_1) at (0,-2.8)  {$u_1$};
    \node[thick,state]         (u_2) at (1.5, 0)  {$u_2$};
    \node[thick,state]         (u_3) at (1.5,-1.7)  {$u_3$};
    \node[circle,draw=black,minimum size=0.26cm,inner sep=0pt,fill=black] (t3) at (1.5,-2.9)  {};
    \node[text width=1cm] at (1.3,-2.9) {$u_T$};
    \path[thick,->, >=stealth] (u_0) edge node [left] {$A$} (u_1);
    \path[thick,->, >=stealth] (u_1) edge node [left] {$B$} (u_2);
    \path[thick,->, >=stealth] (u_2) edge node [right] {$C$} (u_3);
    \path[thick,->, >=stealth] (u_3) edge node [right] {$H$} (t3);
    
\end{tikzpicture}
    \subcaption{}
    \label{fig:deliverytasksequential}
  \end{subfigure}
  \hfill
  \begin{subfigure}[t]{0.33\textwidth}
    \centering
    \begin{tikzpicture}[node distance=cm,on grid,every initial by arrow/.style={ultra thick,->, >=stealth}]
    \node[thick,state,initial above] (u_0) at (0,0) {$u_0$};
    \node[ thick,state]         (u_1) at (-0.7,-1.4)  {$u_1$};
    \node[ thick,state]         (u_2) at (0.7,-1.4)  {$u_2$};
    \node[ thick,state]         (u_3) at (-0.7,-3)  {$u_3$};

    \node[circle,draw=black,minimum size=0.26cm,inner sep=0pt,fill=black] (t3) at (0.9,-3)  {};
    \node[text width=1cm ] at (1.2,-2.7) {$u_T$};

    \path[thick,->, >=stealth] (u_0) edge node [left] {$A$} (u_1);
    \path[thick,->, >=stealth] (u_0) edge node [right] {$B$} (u_2);
    \path[thick,->, >=stealth] (u_1) edge node [left] {$C$} (u_3);
    \path[thick,->, >=stealth] (u_2) edge node [right] {$C$} (u_3);
    \path[thick,->, >=stealth] (u_3) edge node [below] {$H$} (t3);

\end{tikzpicture}
    \subcaption{}
    \label{fig:deliverytaskdisjunction}
  \end{subfigure}
    \hfill
  \begin{subfigure}[t]{0.33\textwidth}
    \centering
    \begin{tikzpicture}[node distance=cm,on grid,every initial by arrow/.style={ultra thick,->, >=stealth}]
    \node[thick,state,initial above] (u_0) at (0,0) {$u_0$};
    \node[ thick,state]         (u_1) at (-0.9,-1.3)  {$u_1$};
    \node[ thick,state]         (u_2) at (0.9,-1.3)  {$u_2$};
    \node[ thick,state]         (u_3) at (-0.9,-3)  {$u_3$};
    \node[ thick,state]         (u_4) at (1,-3)  {$u_4$};

    \node[circle,draw=black,minimum size=0.26cm,inner sep=0pt,fill=black] (t3) at (1,-2)  {};
    \node[text width=1cm ] at (1.75,-2) {$u_T$};

    \path[thick,->, >=stealth] (u_0) edge node [left] {$A$} (u_1);
    \path[thick,->, >=stealth] (u_0) edge node [right] {$B$} (u_2);
    \path[thick,->, >=stealth] (u_1) edge node [left] {$B$} (u_3);
    \path[thick,->, >=stealth] (u_2) edge node [left] {$A$} (u_3);
    \path[thick,->, >=stealth] (u_3) edge node [below] {$C$} (u_4);
    \path[thick,->, >=stealth] (u_4) edge node [left] {$H$} (t3);

\end{tikzpicture}
    \subcaption{}
    \label{fig:deliverytaskcomposite}
  \end{subfigure}
  \caption{Finite state automatons for the Delivery domain (sequential (d), disjunction (e) and composite (f)) tasks.}
 \label{fig:fsas_delivery}
\end{figure*}

The FSA task specifications used in the experiments are fully described in Figure~\ref{fig:fsas_office} (Office environment) and Figure~\ref{fig:fsas_delivery} (Delivery environment). We divide the tasks into three types and their natural language interpretation is as follows:

\begin{itemize}
    \item {\bf Sequential} `Get coffee, then get mail and then go to an office location' (Office domain, Figure~\ref{fig:officetasksequential}) and `go A, then B, then C and then H' (Delivery domain, Figure~\ref{fig:deliverytasksequential}).
    \item {\bf Disjunction} `Get coffee OR get mail, then go to an office location' (Office domain, Figure~\ref{fig:officetaskdisjunction})) and `go to A OR B, then to C, and then to H' (Delivery domain, Figure~\ref{fig:deliverytaskdisjunction}).
    \item {\bf Composite} `Get coffee AND get mail in any order, then go to an office location' (office domain, Figure~\ref{fig:officetaskcomposite})) `go to A AND B in any order, then go to C, then H' (Delivery domain, Figure~\ref{fig:deliverytaskcomposite}).
\end{itemize}
Note that agents must satisfy such tasks in the least possible number of steps.

\end{document}